# A Predictive Model for Kidney Transplant Graft Survival using Machine Learning

Eric S. Pahl[1], W. Nick Street[2], Hans J. Johnson[3] and Alan I. Reed[4]

[1]Health Informatics, University of Iowa, Iowa, USA
[2]Management Sciences, University of Iowa, Iowa, USA
[3]Electrical and Computer Engineering, University of Iowa, Iowa, USA
[4]Organ Transplant Centre, University of Iowa, Iowa, USA

*ABSTRACT*

*Kidney transplantation is the best treatment for end-stage renal failure patients. The predominant method used for kidney quality assessment is the Cox regression-based, kidney donor risk index. A machine learning method may provide improved prediction of transplant outcomes and help decision-making. A popular tree-based machine learning method, random forest, was trained and evaluated with the same data originally used to develop the risk index (70,242 observations from 1995-2005). The random forest successfully predicted an additional 2,148 transplants than the risk index with equal type II error rates of 10%. Predicted results were analyzed with follow-up survival outcomes up to 240 months after transplant using Kaplan-Meier analysis and confirmed that the random forest performed significantly better than the risk index ($p<0.05$). The random forest predicted significantly more successful and longer-surviving transplants than the risk index. Random forests and other machine learning models may improve transplant decisions.*

## 1. INTRODUCTION

There is little research into the methodology regarding how organ transplant stakeholders make decisions, predictions, and assessments of viability and matching for deceased donor kidney transplantation at the time of organ offer and acceptance. Clinical decisions have traditionally relied on the time-tested orthodoxy of data derived from Cox regression-based models to provide statistical relevance to decision making. Recent advances in machine learning (ML) methods provide the opportunity to create highly nonlinear models with complex interactions among variables that may provide superior predictive power. Machine learning usage in medical domains is increasing as demonstrated by successful applications for predicting better utilization of perioperative antibiotics, predicting hospital lengths of stay, and indeed even recently in formulating alternative models for organ distribution and post-transplant implied utilization [1-5]. Despite evidence-based clinical and cost advantages of transplantation, nearly 1 in 5 viable deceased-donor kidneys procured are discarded (~4,000 per year) [6]. Many studies have demonstrated a significant survival benefit for wait-listed patients to accept (or for centers to accept on their behalf) any kidney available for transplant, regardless of the current acceptance metrics [7-10].

Healthcare professionals can better understand the risk of transplantation with models that capture an individual's health state more entirely in the context of a specific prospective organ variables. Our paper uses ML to recreate the Kidney Donor Risk Index (KDRI) to determine if ML can lead to a better predictive model than the Cox regression initially used. The Cox regression employed to develop the KDRI resulted in a piecewise linear formula used both in donor allocation and distribution (recipient acceptance criteria). We will follow the guidelines established by Wei Luo, et al. 2016 to develop and report the machine learning predictive models comparison [11].

The KDRI is commonly adjusted annually and implemented as the derivative metric, the Kidney Donor Profile Index. The KDRI was developed in 2009 and has been the industry and regulatory standard for kidney quality since 2011 [12]. Despite the need and the industry's enthusiasm for the adoption of this metric, the KDRI has many limitations. The KDRI has a 0.600 measure of predictive quality represented by the area under the receiver operating characteristic (ROC) curve (AUC). The KDRI was developed using a Cox proportional hazard regression method. The final KDRI model included 15 variables measured from the donor, resulting in a piecewise linear model that suffers biases when arbitrarily categorized variables are present. For example, in the KDRI model the risk coefficient change based on age over or under 18 and 50; weight over or under 80; creatinine over or under 1.5; etc. [12]. By design, the KDRI model incorporates only variables measured from the donor to assess the risk of transplantation for a recipient.

Further, it would seem logical that predictive measures of transplant outcomes should incorporate recipient variables, like those present in the estimated post-transplant survival score (EPTS) to further improve the modeling. The EPTS was also developed as a piecewise linear model (age over or under 25), using four recipient variables, and is used separately in allocation algorithms. Because EPTS and KDRI were constructed independently of one another, the simultaneous use of these separate models, as in the current allocation system, cannot capture interactions among the donor and recipient variables. A combined model including readily available variables from the donor and recipient, utilizing machine learning, may improve the predictive capabilities for longer-term graft survival.

## 2. METHODS

The methods are described using the clinical setting, prediction problem, data for modelling, and predictive models.

### 2.1. Clinical Setting

The clinical setting and objective for our experiment was to predict kidney transplant outcomes with the data present at the time of organ allocation; the same data a clinician would have or could approximate at the time of organ offer. A clinically adequate predictive model may assist clinical decision-making at the critical time when choosing to accept or decline an organ offered for transplant. We used nation-wide kidney transplantation data from the United Network for Organ Sharing.

## 2.2. Prediction Problem

The prediction problem was the classification of retrospective data with three prognostic binary outcomes: graft failure at 12, 24, and 36 months after transplant were abbreviated as GF12, GF24, and GF36 respectively. The positive (+) outcome for GF12 meant that we observed graft loss within 12 months after transplant and the negative (-) outcome for GF12 meant that graft had not failed at 12-month follow-up. A false positive (type I error) in prediction of graft failure may lead to a declined kidney that would have been successful (missed opportunity). A false negative (type II error) may lead to an accepted kidney that failed within the follow-up period. For this experiment, we highlighted and compared all predictive models' performance at 10% false negative rate because the overall observed graft failure rate is 10% at 12 months follow up. Models with more predicted negative GF outcomes were able to achieve higher utility of available kidneys without increasing risk to patients. We validated the performance of our models using 10-fold cross-validation, whereby we blinded the models to 10% held-out data for testing and reported the test performance only [13].

Table 1 contains descriptions and definitions of predictions; true positive (TP), true negative (TN), false positive (FP), and false negative (FN). The models' predictions are labeled "Predicted GF" and the observed transplant outcomes are "Actual GF." When the Predicted GF and Actual GF are the same, the model predicted correctly. There are two types of model prediction errors; False Positives (Type I errors) happen when a model predicts that the graft would fail but the observed data show success, this is a missed opportunity; False Negatives (Type II errors) happen when a model predicts that the graft would survive but the observed data show failure. False Negatives are the worst prediction errors from a clinical perspective; predicting a success, transplanting, and the graft failing or patient dying.

Table 1. Definitions of prediction outcomes when models were used to predict graft failure (GF) within the months following a kidney transplant.

| Confusion Matrix | Predicted GF | Actual GF | Description |
|---|---|---|---|
| True Positive | Failed (1) | Failed (1) | Model Correctly Predicted a Bad Kidney (avoided a bad kidney, bigger is better) |
| True Negative | Not Failed (0) | Not Failed (0) | Model Correctly Predicted a Good Kidney (bigger is better) |
| False Positive (type I error) | Failed (1) | Not Failed (0) | Model Error Missed a Good Kidney (missed opportunity, smaller is better) |
| False Negative (type II error) | Not Failed (0) | Failed (1) | Model Error Missed a Bad Kidney (graft failure, smaller is better) |

## 2.3. Data for Modelling

We reconstructed, as closely as possible, the original data set used by Rao for the development of KDRI [12]. First, we obtained Standard Transplant Analysis and Research data inclusive of September 29, 1987, through March 31, 2016. The data were filtered by transplantation date within the acceptable date range (01/01/1995 to 12/31/2005) and only included deceased donor kidneys, Initial Data $n_i$, (see data reduction in the first row of Table 2). Excluded in sequence from the analysis were: recipients aged less than 18 years, recipients with a previous transplant, multi-organ transplant recipients, and ABO-incompatible patients, keeping consistent with Rao. We also removed observations with invalid and/or missing data: donor height (<50cm, >213cm), weight (<10kg, >175kg), and creatinine (<0.1mg/dL, >8.0mg/dL). Finally, we removed observations without a valid entry for the KDRI_RAO variable. Table 2 compares the number of removed observations from Rao's study and our study for each missing or invalid variable [12].

Table 2. Observed preprocessing of the UNOS STAR data compared to KDRI development data from kidney transplants 1995-2005.

| Removed Observations | KDRI | Observed |
|---|---|---|
| Initial Data | 92102 | 91996 |
| Pediatric Transplants | 3733 | 3724 |
| Previous Transplants | 13122 | 12390 |
| Multi-Organ Transplants | 1556 | 1850 |
| ABO Incompatible Transplants | 211 | 176 |
| Invalid/Missing Donor Height | 2481 | 1009 |
| Invalid/Missing Donor Weight | 667 | |
| Invalid/Missing Donor Creatinine | 892 | 949 |
| Without "KDRI_RAO" | 0 | 1181 |
| Final Study Sample | 69440 | 70242 |

The predictor variables were the same as the ones present in KDRI and EPTS. Donor variables were those used in the final model of KDRI: age, race, history of hypertension, history of diabetes, serum creatinine, cerebrovascular cause of death, height, weight, donation after cardiac death, hepatitis C virus status, HLA-B and HLA-DR mismatching, en-bloc transplant, and double kidney transplant indicators (two for one), and known cold ischemia time at time of offer. Transplant recipient variables from EPTS score were used in our MLM; recipient age, diabetes, and time on dialysis - notably excluding recipients with prior transplant and multiorgan transplants based on the filter criteria.

We evaluated the predictive performance of the models with an industry standard 10-fold cross-validation approach [13]. The completion of the cross-validation process yielded a ranked list of predicted transplant outcome probabilities. In 10-fold cross-validation, the models are evaluated on external test data that are never used in training. We evaluated the models, using the predicted ranked lists, by generating the ROC curve and calculating the AUC as standard measures of predictive quality and analyzed the confusion matrices with false negative rates at 10%. Additionally, performed Kaplan-Meier survival analysis with survival groups selected by prediction cut-off at a 10% false negative rate.

## 2.4. Predictive Models

In contrast to Rao's Cox proportional hazard regression method, we explored a supervised random forest (RF) ML classification model. The RF algorithm constructs and combines the predictions of thousands of machine-generated decision trees to model the probability of graft failure [14]. The algorithm created each decision tree using a subset of the training data called a bootstrap sample. Each bootstrap sample is balanced by under sampling the majority outcome cases (negatives) such that the resulting ratio was 1:1. Each tree consists of multiple decision nodes constructed by randomly selecting a subset of the predictive variables and choosing the one that maximizes the Gini index, a measure of information gained from the use of each variable. Training continued until "exhaustion": each tree completely fits the training sample. When classifying new examples, all the trees made a prediction, and the output of the RF was the percentage of votes for each outcome, one vote per tree, and the predicted outcome was assigned based on majority vote.

The choice of the RF algorithm allowed for mixed data types (binary, categorical, and numerical) without scaling or significant data modification. The RF training is computationally efficient, is robust to outliers and co-linearities, contains simple tuning parameters, and has demonstrated success for a variety of healthcare data applications [14, 15]. RFs have also demonstrated utility in predicting deceased donor organ transplantation success and offer acceptances in simulated organ allocation models [3, 4].

The contribution of each tree in an RF was similar to getting thousands of opinions based on a professional colleague's background. The clinical implementation of the RF algorithm worked similarly to secondary and tertiary opinions among professionals across the country convening for the treatment of a complicated case. A random subset of all available variables and clinical observations informed the construction of every decision tree and resulted in a unique perspective represented by each tree. The majority vote among the decision trees, was the final prediction of graft failure. The three models tested were: RF using only donor variables from KDRI (RFD), RF using donor variables from KDRI and recipient variables from EPTS (RFDR), and Rao's KDRI. Figure 1 demonstrates the different numbers of trees used for RFD and RFDR models including stratifying and balancing to obtain BSS and the resulting AUC. Increasing the number of trees improved the AUC reached by RFD and RFDR, and both converged between 1000 and 1500 trees. The number of trees was a significant hyperparameter for the RF algorithm. KDRI results do not depend on the number of trees.

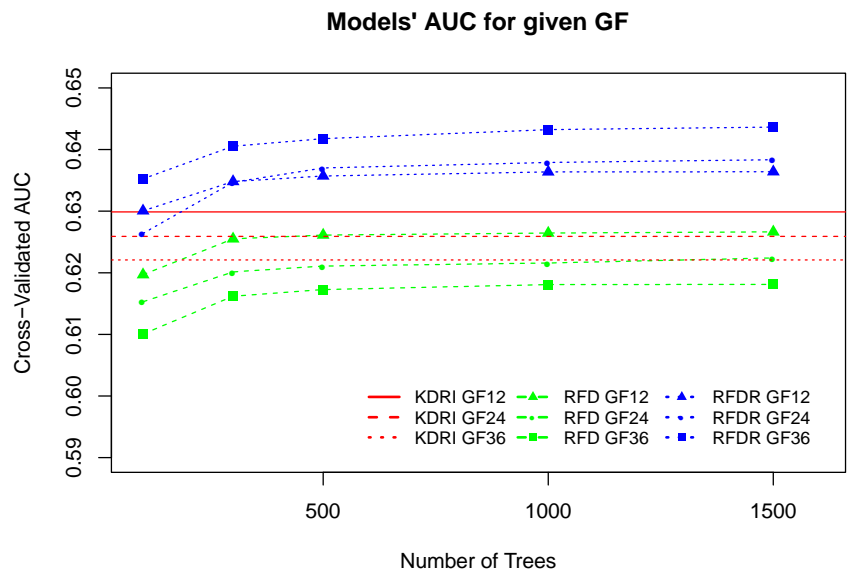

Figure 1. AUC for RF models created with different numbers of trees. KDRI model's AUC performance does not depend on the number of trees.

## 3. RESULTS

RFDR performed better than RFD in all scenarios: adding recipient variables improved the predictive quality of the RFDR model. This is particularly true when predicting longer graft survival outcomes where RFDR performed increasingly better with longer time periods; GF12 (AUC = 0.636), GF24 (AUC = 0.638), GF36 (AUC = 0.644). RFD and KDRI, both models built without recipient variables, did not predict as well as the RFDR model did at the longer graft outcomes.

Figure 2 shows ROC curves for each predictive model under different graft failure considerations. In Figure 2 (a), KDRI (AUC 0.630) performed the same as RFD (AUC 0.627, p = 0.317) and RFDR (AUC 0.636, p = 0.052) at GF12. In Figure 2 (b), KDRI (AUC 0.626) performed the same as RFD (AUC 0.622, p = 0.187) and significantly worse than RFDR (AUC 0.638, p < 0.000) at GF24. In Figure 2 (c), KDRI (AUC 0.622) performed the same as RFD (AUC 0.618, p = 0.096) and significantly worse than RFDR (AUC 0.644, p < 0.000) at GF36.

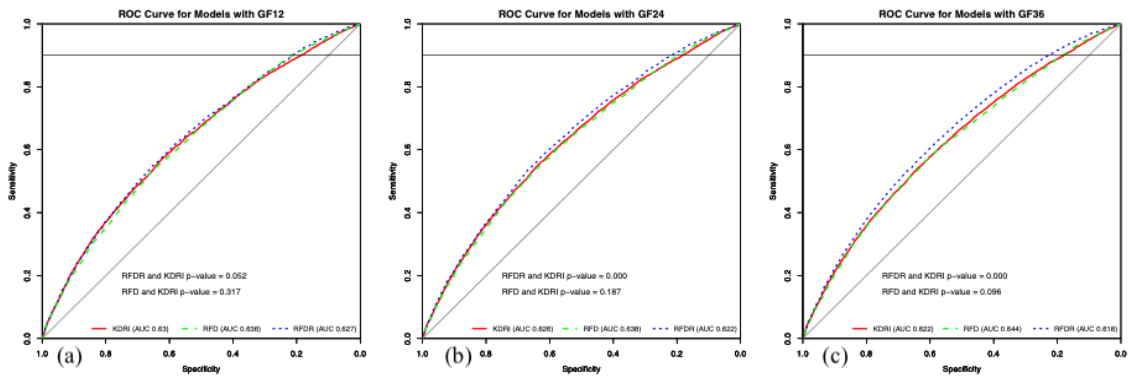

Figure 2. (a-c) ROC curves of three predictive models for different graft failure criteria. P-values were calculated from DeLong's comparison method between ROC curves for RFD and RFDR respectively vs. KDRI are shown. The vertical axis, Sensitivity, at 0.9 is equivalent to the 10% FNR cut-off used for comparing the models in Table 3. The diagonal line represents a ROC curve with an AUC of 0.500.

Table 3 (a) describes the predictions for each model fixed for a 10% FNR at GF 36 months for purposes of direct comparison (Type II error; model predicts success, but graft fails). The FNR cut off maintains that TP and FN are the same for each model. The percentages shown in Table 3 are compared to KDRI. RFD and RFDR performed significantly better than KDRI when predicting which kidney transplant matches would succeed (TN). RFD identified 154 (2%) more successful kidney transplants than KDRI. The RFDR, with additional recipient criteria, identified 2148 (26%) more successful kidney transplant matches than KDRI. Table 3 (b) shows the effect of using different FNR comparison cut-offs for the comparison of KDRI and RFDR at 36GF prediction. The TN, Delta TN, and other counts of additional correct predictions for successful transplants were accumulated through the entirety of the 10-year study period, 1995-2005.

Table 3. (a-b) (a) Comparison of predictions for GF36 with KDRI, RFD, and RFDR models held at 10% FNR cut off. (b) Comparison of predictions for GF36 between KDRI and RFDR models at various FNR.

| Removed Observations | KDRI | Observed |
|---|---|---|
| Initial Data | 92102 | 91996 |
| Pediatric Transplants | 3733 | 3724 |
| Previous Transplants | 13122 | 12390 |
| Multi-Organ Transplants | 1556 | 1850 |
| ABO Incompatible Transplants | 211 | 176 |
| Invalid/Missing Donor Height | 2481 | 1009 |
| Invalid/Missing Donor Weight | 667 | |
| Invalid/Missing Donor Creatinine | 892 | 949 |
| Without "KDRI_RAO" | 0 | 1181 |
| Final Study Sample | 69440 | 70242 |

Figure 3 (a) - (f) shows the Kaplan-Meier survival curves for each model in monthly intervals. Predicted failure (+) and predicted success (-) groups were split based on FN rate (FNR) at 10%. Results show the percentage of patients surviving in each group on the vertical axes and up to 250 months following transplant on the horizontal axes. Figure 3 (a) - (c) show KDRI and RFD directly compared at GF12, GF24, and GF36. Figure 3 (d) - (f) show KDRI and RFDR directly compared at GF12, GF24, and GF36. In all comparisons, FNR was held constant at 10% making all the predicted failure groups (+) statistically similar. RFD survival groups (-) were statistically similar when compared with KDRI survival groups (-) in GF12, GF24, and GF36. RFDR survival groups (-) were statistically significantly better than KDRI survival groups (-) in all GF classifications.

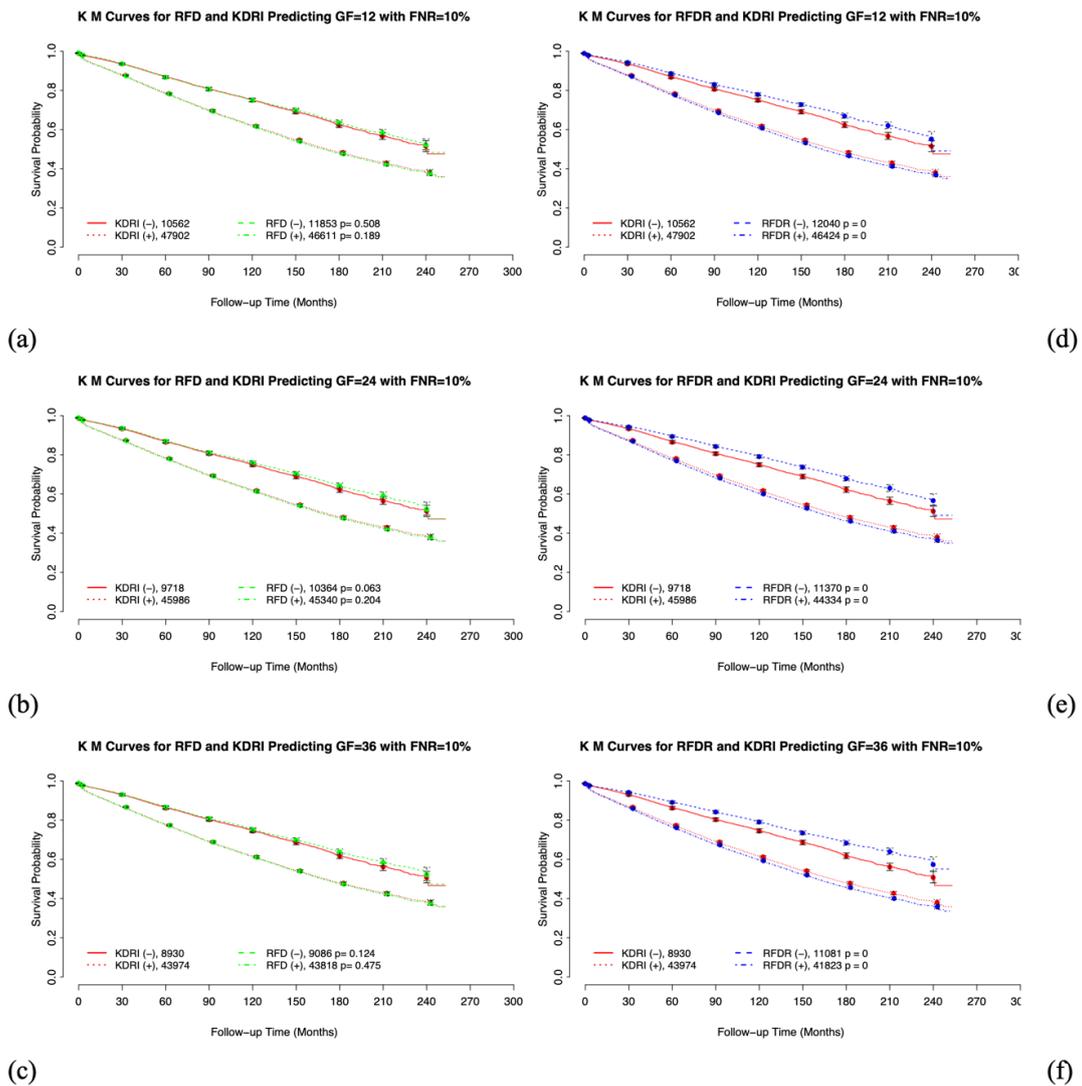

Figure 3. (a-f) Kaplan-Meier (KM) survival analysis comparison for KDRI and RF models using different graft failure criteria to split predicted survival groups for each model. Legends include the label for each line, the population size for that group at time zero, and the p-value calculated from the log-rank test between respective predicted outcome groups (i.e., - / +).

Models with only donor variables, KDRI and RFD, showed decreased longevity predictions whereas RFDR (both donor and recipient variables) better predicted outcomes at GF24 and GF36. The survival observations that trend from the RFDR GF36 (-) group extend past 250 months after transplant and are significantly better than the KDRI (-) group; this trend continues to grow over time. RFDR makes more successful graft predictions for survival at 36 months than KDRI and these grafts survive significantly longer in aggregate.

The final RFDR model reported what variables were shown to be predictive of graft failure, listed in order of importance; Donor Age, Donor Weight, Recipient Time on Dialysis, Organ Cold Ischemia Time, Donor Height, Recipient Age at Transplant, Recipient Age at Waitlisting, and Donor Creatine. This applies specifically to the subpopulation of the high-KDRI donors that has the best prediction and which subpopulation is most difficult to predict. Like the KDRI, it is easier to predict graft failure at the extreme ends of donor quality and more difficult to predict outcomes in the middle. RFDR was able to find more successful outcomes than KDRI, suggesting that the same limitations exist for both models but lesser for RFDR.

## 4. DISCUSSION

Safer clinical decisions informed by ML could empower clinicians to transplant more organs into patients resulting in better outcomes. The margin of improvement by our ML amounted to 2148 additional kidney transplants over 10 years (about 200 per year) that were correctly classified as successful transplants at the same (10% FNR) error rate as KDRI. These results are clinically significant because the tools used by transplant teams will influence life-changing decisions. A clinical team with KDRI and an acceptable 10% graft failure rate at 36 months may be more conservative for kidney offers that would have been successful. The same clinical team with ML may be influenced to be more aggressive for the same kidney offers and capture an added 26% of successful kidney transplants. Deploying ML in a live clinical setting may be an avenue to increase the number of deceased donor kidney transplants without sacrificing patient outcomes.

The inclusion of recipient data improved AUC and longevity prediction; this makes clinical sense and can be refined by adding more data points in the future. Demonstrated improvements in long-term graft survival are associated with increased patient quality-adjusted life years, lower patient health costs, and increased value for all stakeholders [8]. These are the types of decisions at the time of organ offer that can drive real value for the system as opposed to those that optimize only measures of short term success. Studies such as these are extremely timely as professional societies, regulatory agencies and others seek metrics and strategies to drive overall system performance.

Our study design was purposely confined the same data and variables available for KDRI by Rao, et al. 2009, and did not allow us to leverage the full capabilities of ML. Our future work will not have these design constraints. We hypothesize that with the benefit of the additional variables, more recent data, and missing data interpolation, the performance will be greater than what we have demonstrated here. We will expand this research by pursuing more aggressive strategies for optimizing the predictive quality of ML with the inclusion of additional data sources with the ultimate goal of providing real-time decision support to clinicians at the time of organ offer. Increasing the number of transplants has to start with clinicians being able to make the best use of available data at the point of organ offer. This coupled with other larger changes in system dynamics and policy to achieve overall success. ML will allow us to use variables from the recipient, donor, transplant center (administrative, logistical-temporal turndown data) and even behavioral data to predict the transplant outcomes with higher accuracy and clinically relevant predictive quality.

**ACKNOWLEDGEMENTS**

This work was funded in part by the United States National Institutes of Health, National Library of Medicine, award number: R43LM012575. The content is solely the responsibility of the authors and does not necessarily represent the official views of the National Institutes of Health. Eric Pahl received funding to support this research from a fellowship sponsored by the Interdisciplinary Graduate Program in Informatics at the University of Iowa. The datasets analyzed during the current study are available in the UNOS STAR Files repository, [https://optn.transplant.hrsa.gov/data/request-data/].

**AUTHORS**

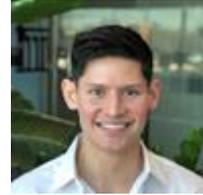

**Eric Pahl**, I am a Ph.D. candidate at the University of Iowa studying Health Informatics an Interdisciplinary Graduate Program in Informatics as part of the Iowa Informatics Initiative. I am also a co-founder of OmniLife (https://getomnilife.com), an early-stage health information technology company developing software to improve the utilization of donated organs and tissues. I am passionate about the development and application of software tools to improve the healthcare services industry. I am pursuing both business and academic opportunities for maximum impact. My ultimate goal is to provide software tools that improve the quality, accessibility, and affordability of world-class healthcare services. I have received Forbes 30 under 30 award for my work in 2018 and since then have received more than $1.75M in federal grant awards from the National Institutes of Health.